\documentclass[conference]{IEEEtran}
\IEEEoverridecommandlockouts
\usepackage{cite}
\usepackage{booktabs}
\usepackage{multirow}
\usepackage{amsmath,amssymb,amsfonts}
\usepackage{algorithmic}
\usepackage{graphicx}
\usepackage{textcomp}
\usepackage{xcolor}
\usepackage{float}
\usepackage{subfig}
\usepackage{tablefootnote}
\usepackage{threeparttable}

\def\BibTeX{{\rm B\kern-.05em{\sc i\kern-.025em b}\kern-.08em
    T\kern-.1667em\lower.7ex\hbox{E}\kern-.125emX}}
\begin{document}

\title{Conference Paper Title*\\
{\footnotesize \textsuperscript{*}Note: Sub-titles are not captured in Xplore and
should not be used}
\thanks{Identify applicable funding agency here. If none, delete this.}
}

\author{\IEEEauthorblockN{1\textsuperscript{st} Given Name Surname}
\IEEEauthorblockA{\textit{dept. name of organization (of Aff.)} \\
\textit{name of organization (of Aff.)}\\
City, Country \\
email address or ORCID}
\and
\IEEEauthorblockN{2\textsuperscript{nd} Given Name Surname}
\IEEEauthorblockA{\textit{dept. name of organization (of Aff.)} \\
\textit{name of organization (of Aff.)}\\
City, Country \\
email address or ORCID}
\and
\IEEEauthorblockN{3\textsuperscript{rd} Given Name Surname}
\IEEEauthorblockA{\textit{dept. name of organization (of Aff.)} \\
\textit{name of organization (of Aff.)}\\
City, Country \\
email address or ORCID}
\and
\IEEEauthorblockN{4\textsuperscript{th} Given Name Surname}
\IEEEauthorblockA{\textit{dept. name of organization (of Aff.)} \\
\textit{name of organization (of Aff.)}\\
City, Country \\
email address or ORCID}
\and
\IEEEauthorblockN{5\textsuperscript{th} Given Name Surname}
\IEEEauthorblockA{\textit{dept. name of organization (of Aff.)} \\
\textit{name of organization (of Aff.)}\\
City, Country \\
email address or ORCID}
\and
\IEEEauthorblockN{6\textsuperscript{th} Given Name Surname}
\IEEEauthorblockA{\textit{dept. name of organization (of Aff.)} \\
\textit{name of organization (of Aff.)}\\
City, Country \\
email address or ORCID}
}

\title{\LARGE \bf
A Robust Illumination-Invariant Camera System for Agricultural Applications
}

\author{Abhisesh Silwal, Tanvir Parhar,  Francisco Yandun and George Kantor}

\maketitle

\begin{abstract}

Object detection and semantic segmentation are two of the most widely adopted deep learning algorithms in agricultural applications. One of the major sources of variability in image quality acquired in the outdoors for such tasks is changing lighting condition that can alter the appearance of the objects or the contents of the entire image. While transfer learning and data augmentation to some extent reduce the need for large amount of data to train deep neural networks, the large variety of cultivars and the lack of shared datasets in agriculture makes wide-scale field deployments difficult. In this paper, we present a high throughput robust active lighting-based camera system that generates consistent images in all lighting conditions. We detail experiments that show the consistency in images quality leading to relatively fewer images to train deep neural networks for the task of object detection. We further present results from field experiment under extreme lighting conditions where images without active lighting significantly lack to provide consistent results. The experimental results show that on average, deep nets for object detection trained on consistent data required nearly \textbf{four} times less data to achieve similar level of accuracy. This proposed work could potentially provide pragmatic solutions to computer vision needs in agriculture.

\end{abstract}

\section{Introduction}
\label{sec::intro}
Identifying fruits, vegetables, and predicting plant diseases early just from images and using that information as a layer in the decision-making process in the production cycle has the potential to revolutionize agricultural industry. Predictive tasks in agriculture such as yield estimation or detection of disease outbreak has always relied on manual workers which quickly becomes a cumbersome task as the size of the orchards and vineyards get larger. Additionally, researchers often describe the computer vision-based approach to agricultural automation as an efficient, low-cost, non-destructive, and scalable process. The combination of above-mentioned advantages and demand from the industry has generated an enormous amount of interest in machine learning-based approach to solving agricultural computer vision problems. 

Detecting fruits and vegetables in images is also a fundamental requirement for any image-based automation task in agricultural fields \cite{silwal2014apple}. It is often regarded as the critical factor for the success of Ag autonomous systems as it provides the ability to perceive information necessary to generate appropriate actions for tasks such as controlling a robot arm for harvesting and autonomous navigation in row crops \cite{silwal2017design} to name few. A typical computer vision pipeline in fruit detection or semantic segmentation is to identify and localize individual fruit/ fruit pixels within the image. Although, this process has received much attention, getting consistent result and generalizing across different field conditions have proven to be extremely difficult. Numerous publications in the past three decades in vision-based agricultural robotics have regarded retrieving visual information from images as one of the major bottle necks to reach commercial maturity \cite{kapach2012computer}. The quest for robust and more generalized fruit detection algorithm has received renewed focus in the research community.

With the advent of deep neural networks in Ag research communities, object detection and semantic segmentation tasks have seen great success in recent history \cite{chu2020deepapple}. In Ag computer vision, most of object detectors \cite{sa2016deepfruits} \cite{kestur2019mangonet} are trained using transfer learning. However, a significant bottleneck exists even for fine-tuning because of lack of data that can be attributed to the vast varieties of cultivars in ag industry. To name few, in the U.S. alone 100 different varieties of apples are grown commercially \cite{uie2} and more than 415 variety of invasive weed species \cite{usda} infest agricultural sites. Sampling a large number of images for each cultivar and its countless sub varieties and maintaining precisely annotated datasets could overwhelm data collection and sharing efforts. Additionally, ag-specific domain expert oversights are often needed when ground truthing images for common tasks such as weed classification, disease detection, etc.  

Images in agricultural sites are taken mainly in outdoor environment with changing lighting conditions which many researchers \cite{gongal2015sensors} often cite as a limiting factor in computer vision application in Ag. Correct image exposure is critical to preserving the perceptual quality of the images \cite{ilstrup2010one}. Regardless of the computer vision algorithm of choice (classical or machine learning- based), features in images are important factors. If the image is under or over-exposed, features and textures can become undetectable that  could negatively impacts the performance of the computer vision algorithms. Thus a robust system to collect images with consistent quality in all lighting condition could provide robust solution.

The contributions of this paper are in the design and evaluation of a high throughput camera system that consistently produces image invariant to changes in environmental lightning. We hypothesize that the consistency in images reduces the amount of data required for training (fine-tuning) deep neural networks  in agricultural applications. This paper is organized in the following way. Section \ref{sec::rel_work} reviews relevant literature in camera systems and the use and implementation of deep learning algorithms for object detection in ag. The hardware design of the camera system is described in Section \ref{sec::methods} which also provides additional details on field experiments and datasets collected and used in this paper. Then, Section \ref{sec::results}  compares the performance of various trained models using active light (AL) and natural light (NL). Finally, Section \ref{sec::conclusions} includes concluding remarks and directions for further research. Throughout this paper, we use the acronyms AL and NL to describe datasets and lighting environments.

\section{Related Work}
\label{sec::rel_work}
\subsection{Object detection in agriculture}
One key aspect of the continuous success in the integration of agriculture and technology is the application of image processing and data analysis techniques for detecting objects within the crops canopy. Then, tasks such as  yield estimation, anomaly evaluation or harvesting could be performed in a more cognizant and efficient way \cite{tian2020computer}. In the past, the images analysis required the use of significant amount of traditional vision computer techniques (e.g., erosion, dilation, contour detection) to detect and/or count objects of interest such as fruits, workers or their tools \cite{syal2014apple, payne2013estimation}. For example, a radial symmetry transform and texture detection techniques were used in \cite{nuske2014automated} to detect and count berries in RGB images, obtaining a $R^2$ correlation of up to 0.95.

Despite the successful results of traditional computer vision techniques in some applications, there are others that are extremely difficult to design/engineer due to the challenging characteristics of the agricultural environments (e.g., changing illumination, blurring) \cite{gongal2015sensors}. In these conditions, deep learning approaches have proven to be an useful tool for image classification and object detection \cite{kamilaris2018deep}. Activities like leaf disease detection, weed/fruit counting, or plant type detection have been robustly performed in different conditions and scenarios \cite{yu2019deep, ferentinos2018deep, vasconez2018toward}. For example, a CNN with a custom architecture was used in \cite{chen2017counting} to count oranges apples in a cluttered scene, obtaining accuracies up to 0.957 and 0.961, respectively. In contrast to using custom architectures, other authors employ known networks (or slightly modified versions) such as Faster-RCNN , Yolov3, Single Shot MultiBox Detector, among others \cite{osorio2020deep, jiang2019real, rahnemoonfar2017deep}.

The proliferation of large scale datasets like \cite{everingham2010pascal}, \cite{lin2014microsoft}, \cite{russakovsky2015imagenet} and  \cite{kuznetsova2018open}  has been a major factor in the recent state of the art results in object detection. But these large scale datasets are generally restricted to generic objects like cars, pets etc. While there is a body of image datasets for agricultural perception tasks, they are generally collected in well curated laboratory conditions \cite{redmon2018yolov3}, \cite{mohanty2016using}, \cite{lobet2017image}, \cite{lobet2013online}. For robust computer vision in field settings, there is need for datasets that collected under more realistic field conditions. And, if robustness to varying lighting conditions needs to be tackled, there is a need for a dataset with that contains images representing all the varying field conditions \cite{sun2017revisiting}.

As computer vision continues to become an important aspect of precision agriculture, to further advance research  a variety of datasets  have been released either through publications or independently. For example, \cite{sa2016deepfruits} consists of 587 bounding boxes for various fruits in RGB images, \cite{altaheri2019date} consists of more than 10000 images for Date fruit yield estimation, \cite{kestur2019mangonet} consists of 49 images with pixel-wise labels for mango segmentation and \cite{bargoti2017deep} contains 3704 images of orchards fruits like apples and almonds. While this is a much anticipated and encouraging direction, the challenge of collecting datasets comparable to the size of \cite{lin2014microsoft} in agricultural cultivar has always remained illusive.

Robotics  or computer vision-based autonomy in agriculture needs to be resilient to  outdoor environment. Collecting data to train deep networks that represents all possible modalities of cultivar in Ag is an onerous challenge. Thus, a camera system agnostic to external lighting conditions, that can generate consistent image data could benefit Ag industry. The dataset collected with the camera system in this study will be  publicly shared.

\subsection{Imaging sensors and systems in agriculture}
A review article by \cite{gongal2015sensors} shows that the most popular choice of sensors for fruit detection is mostly color cameras. Nearly two dozen cited articles in \cite{gongal2015sensors} use color cameras in various fruit detection tasks. Other choice of sensors include hyper-spectral, thermal, monochrome camera and in some occasions laser range finders. Remarkably, all research work using color camera in \cite{gongal2015sensors} acknowledge variation in lighting as the limiting factor. An alternative solution often used by researchers to control lighting environments is to use a mobile structure to completely block sun light to image the canopy \cite{botterill2017robot} \cite{silwal2014apple}. In a recent work, \cite{pothen2016automated} used high resolution stereo sensor with flash to predict yield in vineyards from image-based counting for grapes. The use of flash imagery in this study generated images with uniform white balance and had minimal effects from natural illumination. Our design of the camera system in this paper is motivated by this work. To our best knowledge, the affect of consistent image quality on the size of training data for deep neural networks in agricultural has not been previously reported.     
\section{Methods}
\label{sec::methods}
A camera essentially converts incident irradiant light energy into discrete pixel values. The conversion of the irradiant energy to brightness levels in pixel includes transforming the incident photons to electrical charges and then quantize the analog signal to digital data. Eqn.(\ref{equation::pixel}) from \cite{pillman2011camera} shows the expression of the signal received by a pixel $P$ .
\begin{align}
\label{equation::pixel}
P = T K \frac{l^2}{N^2}\int\frac{E_\lambda R\left(\lambda\right)}{\frac{hc}{\lambda}}Q(\lambda) d\lambda  
\end{align}
where, $T$ is the exposure time, $K$ is a normalization constant, $l$ is the pixel pitch, $N$ is  the  f/number  of  the lens and $\lambda$ is the wavelength. The terms $Q(\lambda)$, $E_{\lambda}$ and $R(\lambda)$ are the spectral quantum efficiency, spectral power distribution and spectral reflectance respectively, which are dependent on the wavelength of the incident light. The remaining values $h$ and $c$ are the Plank's constant and the velocity of light. This analog value is then quantized to be represented into digital formats.

Without flash or active lighting, the terms $E_{\lambda}$ and $R(\lambda)$ inside the integral of Eqn.(\ref{equation::pixel}) are variables that depend on the intensity of natural light captured by the sensor. As natural light intensity (brightness) and wavelength (color) change drastically throughout the day, the change in the irradiant energy entering the camera sensor could alter the perceptual quality of object of interest in the image. This unpredictable change in the quality of the images taken in outdoors increases the variability in the image as a data point in computer vision applications. However, with active lighting, $E_\lambda$ and $R(\lambda)$ in Eqn.(\ref{equation::pixel}) become constants just leaving exposure time $T$, digital gain, and aperture to control image exposure.

\subsection{Hardware Design}
\label{sec::hardware}
\begin{figure}[t!]
   \centering
   \subfloat[]{
        \label{Fig::camera_cad}         
        \includegraphics[width=0.35\linewidth]{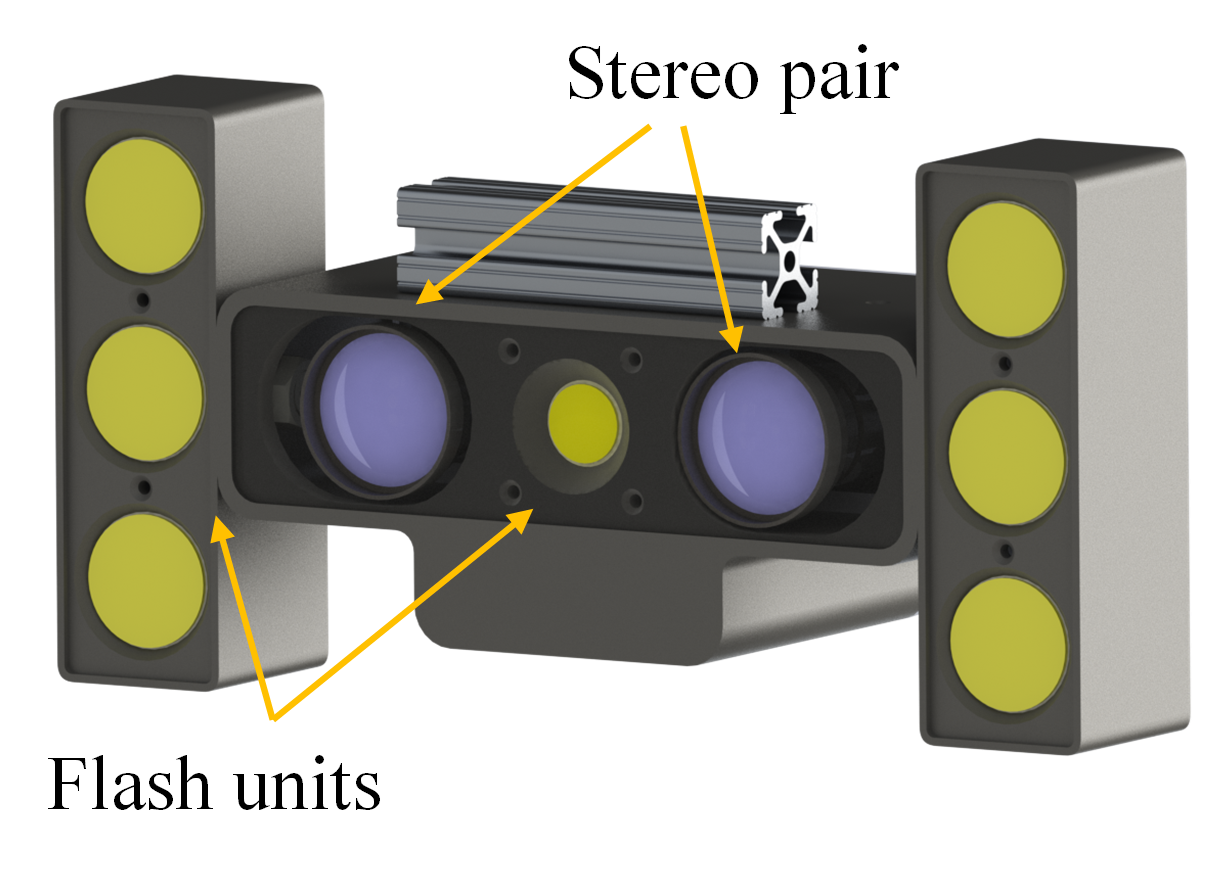}}
   \subfloat[]{
        \label{Fig::hardware_trigger}         
        \includegraphics[width=0.5\linewidth]{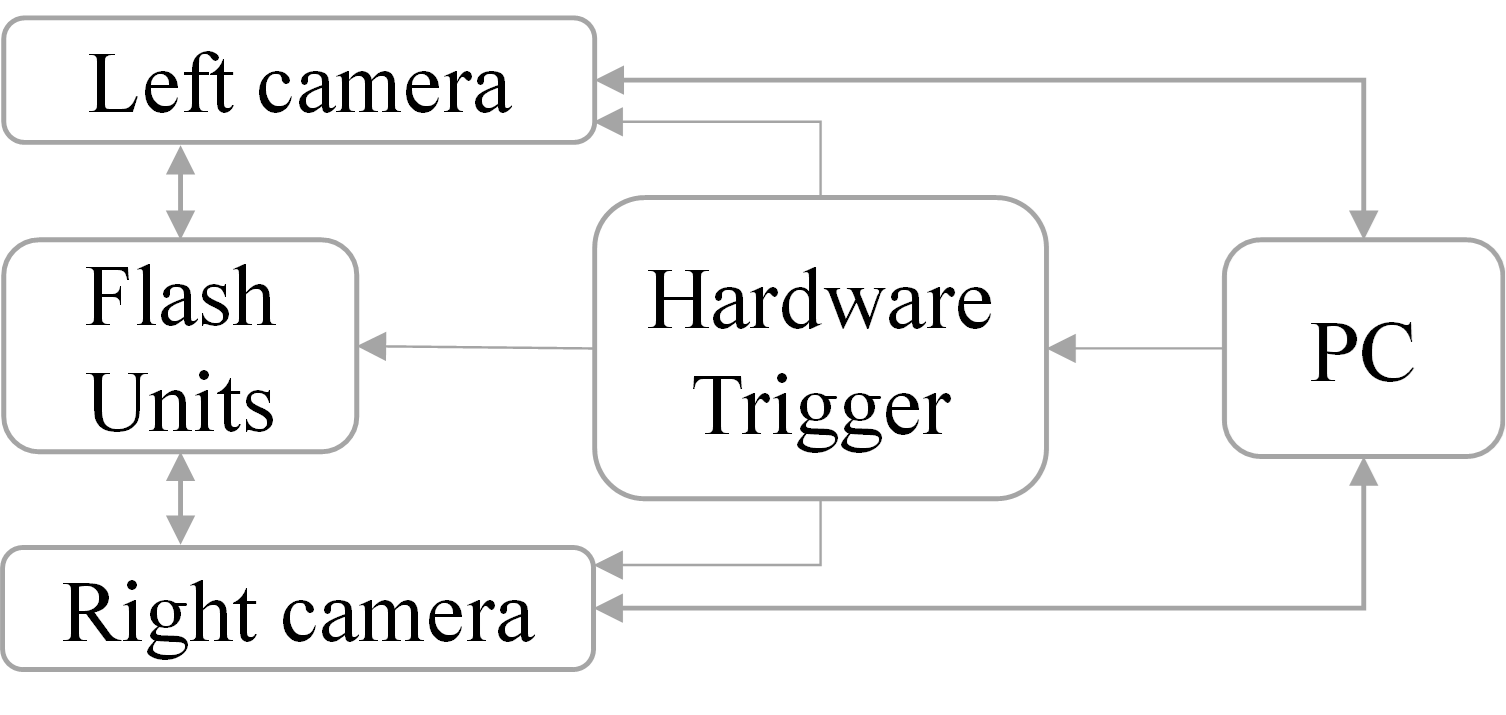}}
   \caption{Camera system (a). Image and flash synchronization flowchart (b)}
   \label{Fig::camera_hw}                
\end{figure}

The proposed camera system shown in Figure \ref{Fig::camera_hw} is a custom-built rugged stereo pair. This field prototype houses two industrial color cameras with spatial resolution of 2048x1536 pixels and a stereo baseline of 110 mm. The featured design is modular and enables quick changes in the orientation and quantity of flash units as required. As mentioned in Section \ref{sec::intro}, the use of active lighting potentially reduces variation in image data. However, in practice the generation of such images is only possible when the flash duration is tightly synchronized with image acquisition pipeline. To achieve such synchronization, an external hardware trigger was necessary to send a control signal to all the cameras and flash units simultaneously. The synchronous signal was generated using a micro-controller on-board connected to the host PC. 

The active lighting system described here uses a high-power Chip on Board (COB) LED lights (1.2 K Watts total) that floods the scene with a bright pulse of white light with color temperature of 5600K. The use of the high-power flash allowed us to set the digital gain of the camera sensor to zero dB, further reducing variables in the control of image exposure while avoiding inducing digital noise in the acquired images. In all experiments, the aperture of the lens was kept fixed at the minimum value of f/2.4 that left exposure time $T$ as the only tuneable variable. As the total number of variables in Eqn.(\ref{equation::pixel}) decreases, better control of image exposure is achievable. Additional hardware details are listed in Table \ref{table::hardware_specs}.

\begin{table}[]
\centering
\caption{Camera hardware components and specifications }
\label{table::hardware_specs}
\resizebox{\columnwidth}{!}{ %
\begin{tabular}{ll}
\toprule
Hardware          & Specifications                                                       \\ \midrule
Camera            & PointGrey CM3, 3.2 MP, Color,   global shutter                       \\
Lens              & 3.5mm f/2.4, 89ºx73.8ºx101.7º FoV,   manual aperture and focus       \\
Flash             & 100 Watts side LEDx6, 500 Watts center LEDx1. 5600K color temp \\
Flash Trigger     & 200 - 250 µs                                                         \\
Shutter Speed     & as low as 11 µs                                                      \\
Acquisition Speed & 1 to 20 Hz synchronized stereo pair               \\ \bottomrule                
\end{tabular}
}
\end{table}

\subsection{Imaging in the outdoors}
\label{sec::imaging}

\begin{figure}[h]
   \centering

   \subfloat[]{%
        \label{Fig::al_multi}         
        \includegraphics[clip,width=0.8\columnwidth]{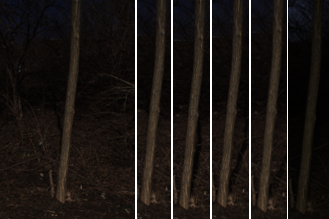}}\\
   \subfloat[]{%
        \label{Fig::dl_multi}         
        \includegraphics[clip,width=0.8\columnwidth]{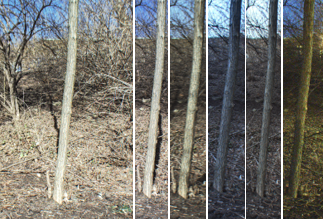}}
   \caption{Image of an outdoor scene with (a) and without flash (b). The snippets next to the main image shows image formation at various times of the day.}
   \label{Fig::multi_exposure}                
\end{figure}

The aim of the first experiment under this section is to quantify the quality of images taken in an outdoor setting with and without active lighting. In this experiment, the camera system described in Section \ref{sec::hardware} imaged an outdoor scene (Figure \ref{Fig::multi_exposure}) every twenty-minute interval from noon to sunset.  First, an image of the scene was captured using the camera auto exposure setting and was immediately followed by the active light imaging. The camera remained in a fixed position throughout the experiment to capture the exact scene for accurate comparison. Additionally, a light meter was used to manually measure the luminance of the scene for all the intervals. For the non-active lighting images, auto exposed and HDR image sets were also collected for comparison. 

The Structural Similarity (SSIM) index proposed in \cite{wang2004image} and defined by Eqn.(\ref{equation:ssim}), and the Peak Signal to Noise Ratio (PSNR) are two of the metrics used to quantify the quality of the captured images in this paper. The SSIM is a quality assessment index which is a multiplicative combination of luminance, contrast, and structural terms. Where as PSNR computed the peak signal to noise ratio in images. Both SSIM and PSNR used the first  image taken at noon as reference. The results of this experiment are detailed in Section \ref{sec::results}.
\begin{equation}
\label{equation:ssim}
    SSIM(x ,y) = f\left(l\left(\mathbf{x},\mathbf{y}\right),c\left(\mathbf{x},\mathbf{y}\right),s\left(\mathbf{x},\mathbf{y}\right)\right)
\end{equation}
With, 
\begin{align*}
l\left(\mathbf{x},\mathbf{y}\right)=\frac{2\left(1+R\right)}{1+\left(1+R\right)^2+\frac{C_1}{\mu_x^2}} \\ 
c\left(\mathbf{x},\mathbf{y}\right)=\frac{2\sigma_x\sigma_y+C_2}{\sigma_x^2+\sigma_y^2+C_2},  \\ 
s\left(\mathbf{x},\mathbf{y}\right)=\frac{\sigma_{xy}+C_3}{\sigma_x\sigma_y+C_3}\\ 
\end{align*}
where $\mu_x,\mu_y, \sigma_x, \sigma_y,$ and $\sigma_{xy}$ are the local means, standard deviations, and image cross-covariance, and $C_1$, $C_2$, $C_3$ are constants. 

The second experiment under this section describes the details of the camera system usage in field. Concretely, it was employed in a commercial apple orchard to generate the dataset used in this paper. Similarly to the first experiment, the apple images were collected with and without the flash. The camera system was mounted on an orchard platform that was manually driven in-between the row-space at approximately 3 mph (0.4 m/s) speed while imaging the tree canopies. The platform was driven twice in the same row to image the fruit with and without active lighting.  

To demonstrate that uniformly exposed images (i.e., consistent in quality) allows a competent training requiring fewer samples, we generated two types of training and testing datasets. The first correspond to images acquired with our active lightning camera system, while the latter comprises images captured in natural lighting. Furthermore, each of the training datasets were divided in three  sub-categories according to the number images: small, medium and large, which contain 20, 40 and 94 samples, respectively. Although the number of images seem relatively small, the field of view of the camera was large enough to capture significant sections of the tree canopies, thus having a larger number of fruit counts per images. The purpose of having datasets with different sizes was not only to compare the performance between NL and AL images at different scales but also to quantify the difference in training samples required to achieve similar performance (See Section \ref{sec::results}). For testing we also included images acquired with the sun facing directly to the camera as an extreme condition for the active and natural lightning (AL Extreme and NL Extreme). Table \ref{table::dataset_details} summarizes the details of all the datasets and also include the total number of fruits, which was counted manually.

The results were validated across four test datasets using four commonly cited deep neural networks in agricultural applications: Faster-RCNN \cite{ren2015faster}, Faster-RCNN + ResNet-50, Single Shot Detector \cite{liu2016ssd}, and YOLO V3 \cite{redmon2018yolov3}. The summary of the parameters used to finetune the networks is listed in Table \ref{table::nnets}  and Table \ref{table::detector_table} summarizes the network performance for each case.

\begin{table}[]
\centering
\caption{Training and test datasets description}
\label{table::dataset_details}
\resizebox{\columnwidth}{!}{%
\begin{tabular}{@{}ccclll@{}}
\toprule
\multicolumn{3}{c}{\textbf{Training Dataset}} & \multicolumn{3}{c}{\textbf{Testing Dataset}}                                                                                                                            \\
\textbf{Name}  & \textbf{No. of Images} & \textbf{Fruit Count} & \multicolumn{1}{c}{\textbf{Name}} & \multicolumn{1}{c}{\textbf{No. of Images}} & \multicolumn{1}{c}{\textbf{Fruit Count}} \\ \midrule
AL Small       & 20                     & 813  & \multicolumn{1}{c}{AL Test}                            & \multicolumn{1}{c}{51}                                          & \multicolumn{1}{c}{1974}                                          \\
AL Medium      & 40                     & 1564                     & \multicolumn{1}{c}{NL Test}                            & \multicolumn{1}{c}{51}                                          & \multicolumn{1}{c}{2272}                                          \\
AL Large       & 94                     & 3218                     & AL Extreme                                            & \multicolumn{1}{c}{20}                                                             & \multicolumn{1}{c}{607}                                                               \\
NL Small       & 20                     & 800                      & NL Extreme                                            & \multicolumn{1}{c}{20}                                                             & \multicolumn{1}{c}{697}                                                               \\
NL Medium      & 40                     & 1695                     &                                                        &                                                                 &                                                                   \\
NL Large       & 94                     & 3315                     &                                                        &                                                                 &                       \\\bottomrule                                 
\end{tabular}
}
\end{table}

\begin{table}[]
\centering
\caption{Training parameters and values.}
\label{table::nnets}
\resizebox{\columnwidth}{!}{%
\begin{tabular}{ccccc}
\toprule
\multicolumn{1}{l}{Network Name} & \multicolumn{1}{l}{Input size} & \multicolumn{1}{l}{Learning rate} & \multicolumn{1}{l}{Epochs} & \multicolumn{1}{l}{Augmentation} \\\midrule
YoloV3 - Darknet-53                           & 512x512                        & 0.001                             & 250                        & HF, S, R                         \\
Faster-RCNN - VGG16                   & 1024x768                       & 0.001                             & 300                        & HF, S, R                         \\
Faster-RCNN - ResNet50                    & 1024x768                       & 0.001                             & 300                        & HF, S, R                         \\
SSD - Inception V2                              & 300x300                        & 0.004                             & 250                        & HF, S, R     \\ \bottomrule                   
\end{tabular}}\\
    \begin{tablenotes}
      \small
      \item HF = Horizontal Flip; S = Scaling; R = Random Crop
    \end{tablenotes}
\end{table}

\section{Results and discussion}
\label{sec::results}
\begin{figure}[h]
   \centering

   \subfloat[]{%
        \label{Fig::graph_with_no_flash}         
        \includegraphics[clip,width=0.9\columnwidth]{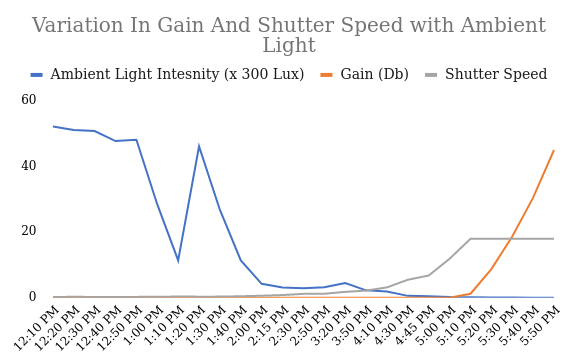}}\\
   \subfloat[]{%
        \label{Fig::graph_flash_SSIM}         
        \includegraphics[clip,width=0.9\columnwidth]{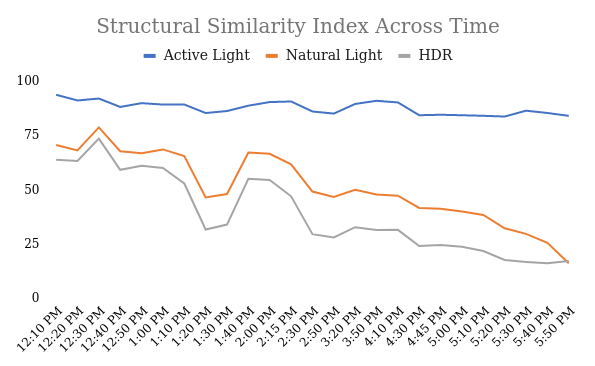}}\\
    \subfloat[]{%
        \label{Fig::graph_flash_PSNR}         
        \includegraphics[clip,width=0.9\columnwidth]{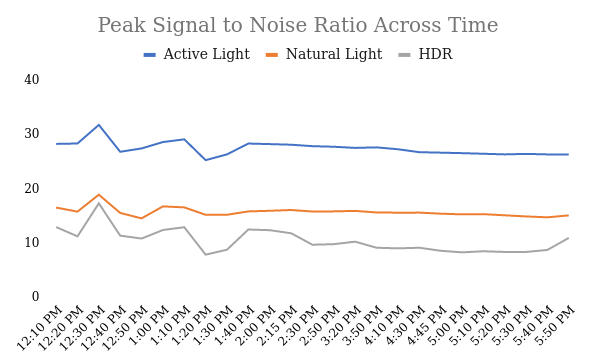}}
   \caption{(a) Change in the intensity of light in a randomly selected day and its effect in the exposure variables (b) Time series plot of SSIM and (c) Time series plot of PSNR of the AL, NL, and HDR images.}
   \label{Fig::image_outdoors}                
\end{figure}

The variation in light intensity from the first experiment described in Section \ref{sec::imaging} is shown in Figure \ref{Fig::graph_with_no_flash}. As various natural factors affected the intensity of the light, the amount of natural light entering the sensor also changes and the camera generates image with different exposures. This change in the image luminosity and contrast captured by the SSIM index is shown in Figure \ref{Fig::graph_flash_SSIM}. For the NL and HDR images, the SSIM quality index shows small fluctuations in time intervals closer to noon (12:01 PM – 1:00 PM, Figure \ref{Fig::graph_flash_SSIM}) and relatively higher SSIM value (approximately 70). This closeness in the SSIM index resembles closeness in quality as the brightness of the sun remained relatively constant. However, beyond this point as more significant intensity changes occurred, image quality between the first image taken at noon and the newly acquired images differed drastically. The image taken at the end of the day (6:00 PM, also seen in Figure \ref{Fig::dl_multi}) shows a high degree of quality difference compared to the images acquired at noon. 

On the other hand, SSIM index of the AL images show not only consistency but higher SSIM value throughout the entire day. Additionally, the PSNR value of AL images are also significantly larger to both NL and HDR images, as depicted in Figure \ref{Fig::graph_flash_PSNR}. As shutter speed was the only variable required to tune exposure, (see Section 3.1), the high-powered flash enables us to set shutter speed to extremely low shutter durations (see Table \ref{table::hardware_specs}). Consequently, at extreme low shutter speeds the number of photons hitting the camera sensors are also very low. Without flash, the images taken under these setting would generate pitch dark images. However, a proper synchronization of flash duration and image acquisition makes the projected light from the flash as the dominating source of illumination. This reflected light that enters back into the camera sensor essentially keeps $E_{\lambda}$ and $R(\lambda)$ constants in Eqn.(\ref{equation::pixel}). Thus, providing consistent quality in all lighting conditions. The outcomes of validating trained deep neural networks on AL and NL test datasets in present in Table \ref{table::detector_table}.

\begin{table}[]
\centering
\caption{Deep Network performance on AL and NL test datasets (AP@0.5 IoU).}
\resizebox{\columnwidth}{!}{%
\begin{tabular}{ccccccc}
\toprule
\label{table::detector_table}
Deep Network  & AL Small       & NL Small & AL Medium      & NL Medium & AL Large       & NL Large \\
YoloV3        & \textbf{0.814} & 0.723    & \textbf{0.806} & 0.756     & \textbf{0.842} & 0.766    \\
Faster-RCNN  & \textbf{0.809} & 0.630    & \textbf{0.810} & 0.714     & \textbf{0.836} & 0.714    \\
Faster-RCNN-RasNet50 & \textbf{0.801} & 0.623    & \textbf{0.813} & 0.699     & \textbf{0.829} & 0.701    \\
SSD           & \textbf{0.624} & 0.515    & \textbf{0.690} & 0.524     & \textbf{0.725} & 0.542   \\ \bottomrule
\end{tabular}
}
\end{table}

\begin{figure}[]
\begin{center}
   \includegraphics[width=0.95\linewidth]{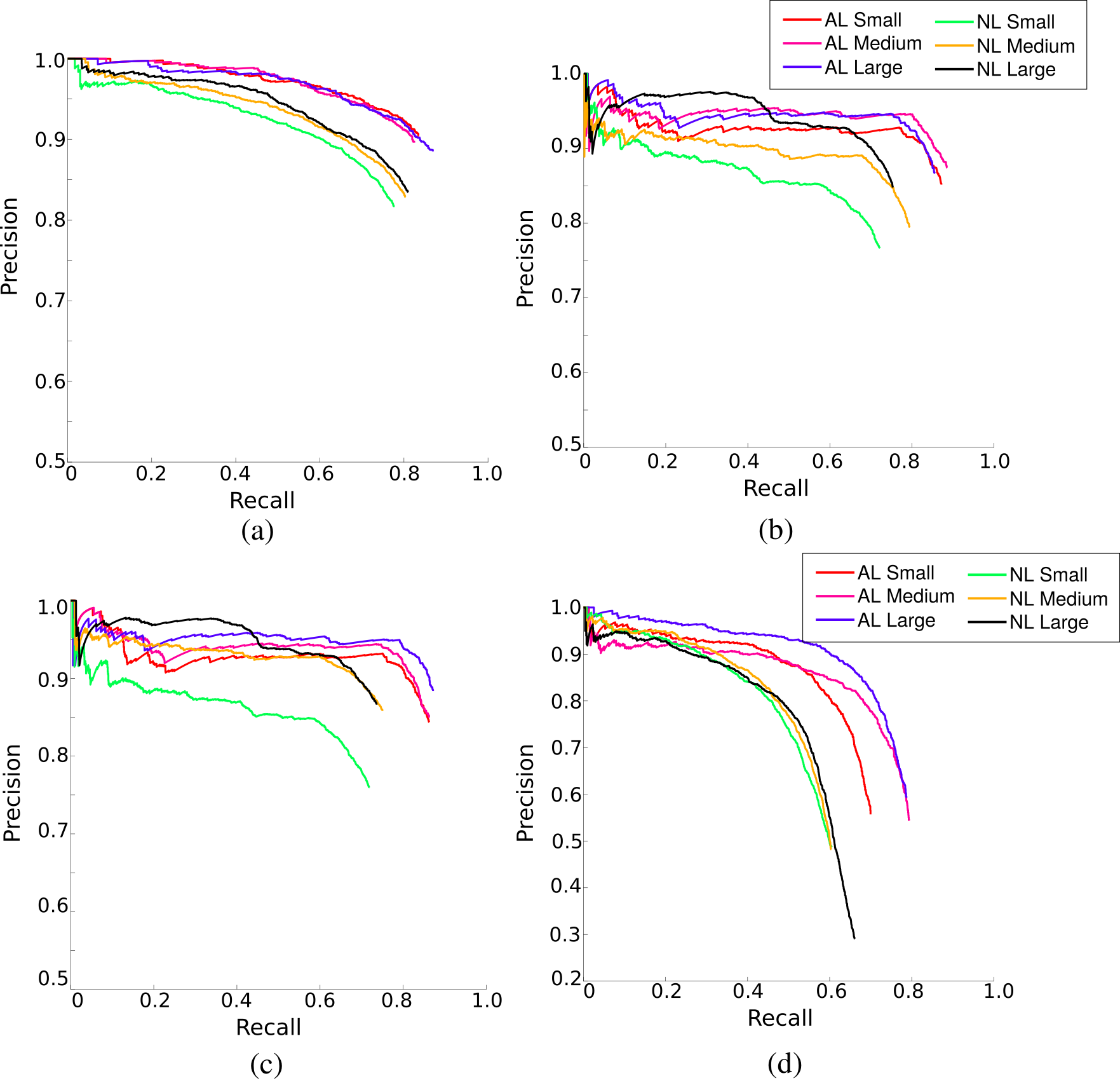}
\end{center}
   \caption{Precision - Recall curves for (a) YOLOv3 (b) Faster-RCNN (c) Faster-RCNN-RasNet50, and (d) SSD}
\label{Fig::pr_curves}
\end{figure}

For brevity, Table \ref{table::detector_table} only shows the average precision (AP@0.5 IoU) as a performance measurement unit for object detection. In all instances, the object detector trained in AL datasets outperformed all their counterparts trained in the NL dataset, for each sizing category. Although the AP on both NL and AL datasets increase with more data, the trend of AP increase in AL dataset has less rate of change, showing consistency for all cases. Additionally, the AP of AL Small trained model for YOLOv3, Faster-RCNN+ ResNet 50 and SSD is close to the AP of NL Large trained model. This shows that the AL Small models have learned to make better inference from relatively less data. Thus, we attribute these observations from Table \ref{table::detector_table} to the quality/consistency of images acquired with active lighting. 

\begin{figure}[h]
   \centering
   \subfloat[]{%
        \label{Fig::al_flair}         
        \includegraphics[clip,width=0.8\columnwidth]{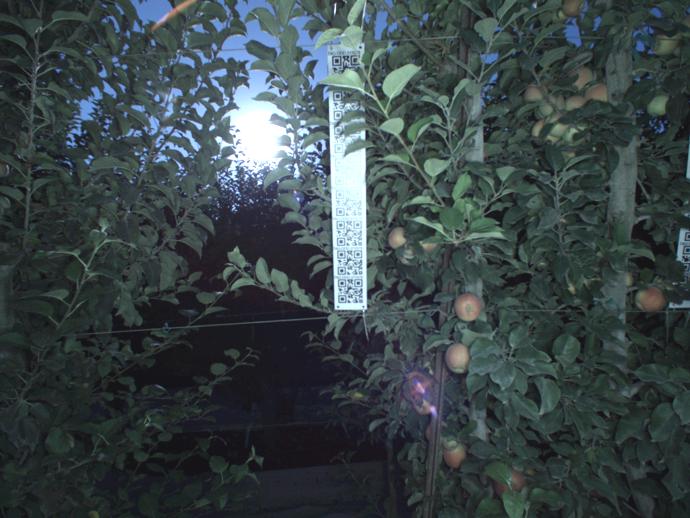}}\\
   \subfloat[]{%
        \label{Fig::dl_flair}         
        \includegraphics[clip,width=0.8\columnwidth]{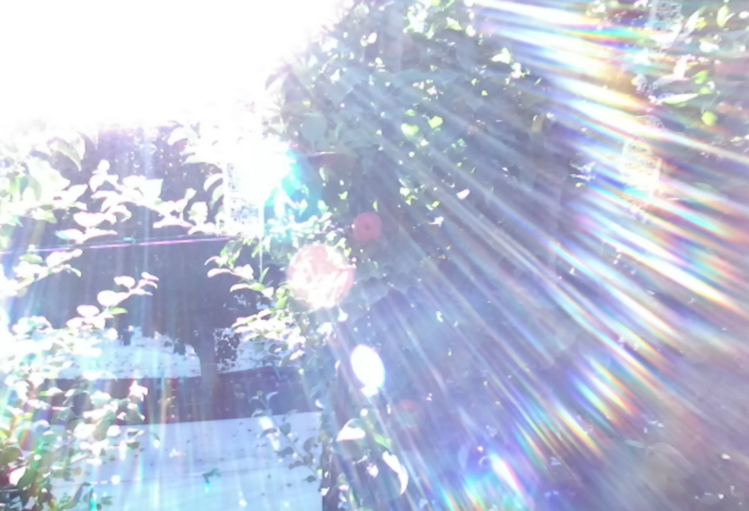}}
   \caption{Sample test images from AL Extreme (a) and NL Extreme (b) datasets with sun flairs.}
   \label{Fig::sun_flair}                
\end{figure}

In the AL and NL Extreme datasets (sun in the background, Figure \ref{Fig::sun_flair}), we further evaluate the robustness of the camera system in extreme situations. Although this situation is labeled as extreme test, sun flairs in images are common occurrences while imagining in the outdoor. Figure \ref{Fig::sun_flair} shows one such situation during the data collection process. Qualitative observation of Figure \ref{Fig::al_flair} shows that the image with active lighting has almost no affect from sun in the background. Where as non active light image (Figure \ref{Fig::dl_flair}) has over-saturated and most of the apples in the scene are not detectable. Faster-RCNN achieved AP of 0.71 on the AL Extreme dataset with the model in AL Large. Noticeably this detection accuracy is close to the detection accuracy of Faster-RCNN on the AL Test dataset (Table \ref{table::detector_table}). On the other hand, because the images were overwhelmed with sun flairs, and perceptual information on the fruit pixels were lost, Faster-RCNN model trained on the NL Large dataset detected negligible amount of fruits, in the NL Extreme dataset counterparts.     

\begin{figure}[h]
   \centering
   \subfloat[]{%
        \label{Fig::al_flair}         
        \includegraphics[clip,width=0.8\columnwidth]{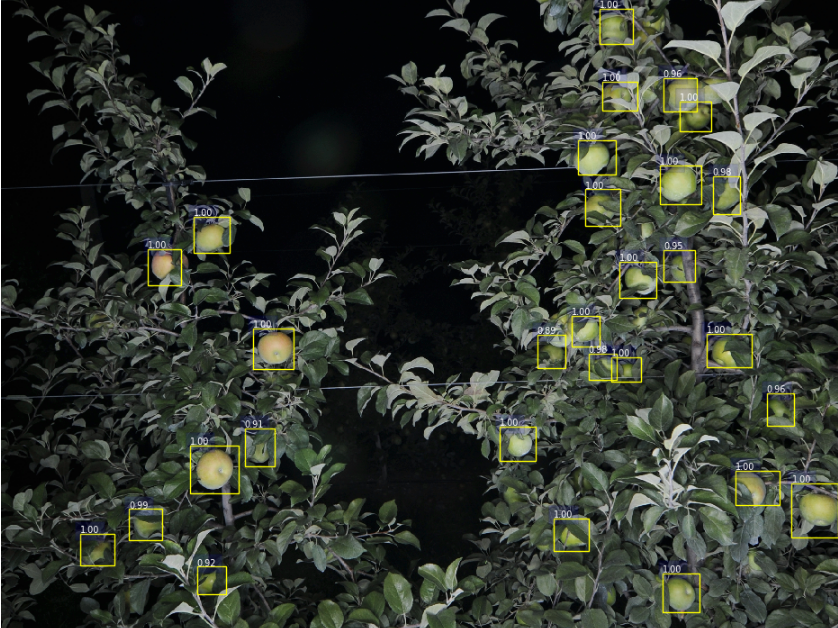}}\\
   \subfloat[]{%
        \label{Fig::dl_flair}         
        \includegraphics[clip,width=0.8\columnwidth]{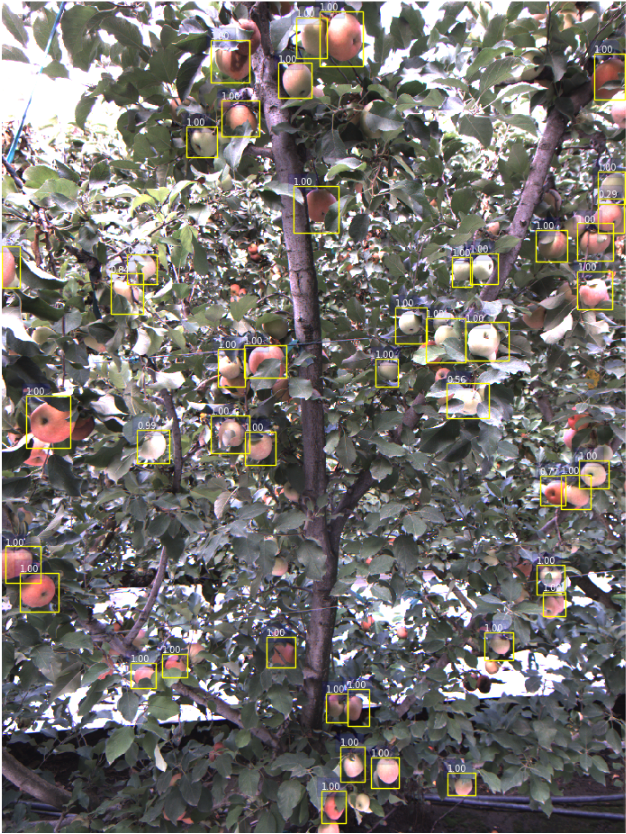}}
   \caption{Sample detection results using Faster-RCNN for AL (a) and NL (b) image.}
   \label{Fig::AL_DL_samples}                
\end{figure}

In addition to consistent quality image and in-variance to external lighting, the proposed camera system offers additional advantages to conventional imaging systems. These advantages are summarized in the following points: 
\begin{itemize}
  \item Background subtraction: The light intensity from the flash attenuates following the principle of inverse square law \cite{voudoukis2017inverse}. This phenomenon is inherently responsible for making the background darker. This effect can also be seen in Figure \ref{Fig::al_flair}. The tree canopy with apples and foliage in the foreground is significantly closer than the flash compared to the adjacent row in the background. The intensity of the flash that reaches the background significantly attenuates and makes it appear relatively darker. However, such affect is not possible in regular images Figure \ref{Fig::dl_flair}. We believe that the effect of darker background could potentially facilitate and increase accuracy in image-based yield estimation of crops as fruit from the adjacent rows are inherently not visible. Further, regular computer vision algorithm for segmentation, object detection, stereo correspondence and others could also benefit.   
  \item No motion blur: Motion blur can occur in images when the camera or the scene or both are in motion. This usually result in dis-figuration of the object and decreases the sharpness of the image. With the use of flash, time required to acquire individual image drastically decrease which consequently  reduces the chances of getting motion blur. This could potentially enable robots to reliably use active lighting camera in outdoors for various tasks such as autonomous navigation and visual servoing. 
  \item Color Consistency: With the light from the flash as the dominating source for illumination, the terms $E_{\lambda}$ and $R(\lambda)$ in Eqn.(\ref{equation::pixel}) become constants, hence the repeatably of generating consistently exposure images get better. The combination of high color rendering index ( which is $\geqslant$ 95\% for the LED flashes) and uniform exposure, render images with high color accuracy.
  \item High throughput: Synchronization of stereo images is a crucial requirement for stereo correspondence. Without precise image synchronization, correspondence between left and right stereo images become a difficult challenge. The stereo pairs used in our camera system are precisely triggered and synchronized with the flash. In field trials, our system could provide synchronized stereo pairs up to 20 Hz. 
  
\end{itemize}

\section{Conclusions}
\label{sec::conclusions}
Although in theory, with additional training images in the NL dataset, similar or higher detection accuracy could be achieved in fruit detection . However, in this paper we took a different approach and focused on the consistency of data rather than size. The focus of attention here was to accurately control image exposure time to generate consistent quality of images in all lighting conditions. With the use of flash as external lighting source, we  demonstrated how the variables in camera exposure can be strategically reduced. Our method showed that it is possible to collect uniformly exposed images with minimum to no effects from the surrounding natural lights. This consistency in image quality allowed us to train deep neural networks with significantly less data while achieving comparable results to networks trained on larger dataset with regular images. Thus the benefit of requiring less data coupled with several advantages including background subtraction, no motion blur, color consistency, and high throughput, our proposed design could provide pragmatic solution to computer vision needs in agriculture. 

Although, the use of active light generated consistently exposed images in this study, the high power of the flash and exponential attenuation of its intensity limits the work range of this camera system. In all experiments described in this paper, the camera was placed between 0.5 to 1.5 meters distance from the canopy. Beyond this range, the images could appear too dark or over exposed. By increasing shutter duration we could still get good quality images in these range, but the resilience to outdoor illuminance could get affected. However, modern vineyards and orchards grow crops in rows with uniform row separation and tree canopies trained in formal architectures to facilitate manual and mechanical operations. This trend towards almost two dimensional canopy structure  provides ideal test environment for our camera system. Currently, the shutter speed during       

Currently in all of our experiments, the shutter duration is intuitively selected based on qualitative observation of foreground highlight vs. background suppression. Although, this approach suffices the need for data consistency  per site, temporal consistency in data could be achieved with auto shutter duration estimation. Our future work will comprise of this task of automating a calibration process to match quality between temporal datasets. 

{\small
\bibliographystyle{IEEEtran} 
\bibliography{main}
}

\end{document}